\documentclass[letterpaper, 10 pt, conference]{ieeeconf} 
\IEEEoverridecommandlockouts                             
\usepackage{color}
\usepackage{graphicx} 
\usepackage{mathtools} 
\usepackage{siunitx}
\usepackage{hyperref}

\usepackage[dvipsnames]{xcolor}
\usepackage{color}

\newcommand{\margaret}[1]{\textcolor{black}{#1}}
\newcommand{\alex}[1]{\textcolor{black}{#1}}
\newcommand{\andrew}[1]{\textcolor{black}{#1}}
\newcommand{\comments}[1]{}

\title{\LARGE \bf Distributed Sensor Networks Deployed Using Soft Growing Robots}

\author{Alexander M. Gruebele$^{*}$, Andrew C. Zerbe$^{*}$, Margaret M. Coad, Allison M. Okamura, and Mark R. Cutkosky
\thanks{*These authors contributed equally to this work.}%
\thanks{The authors are with the Department of Mechanical Engineering, Stanford University, Stanford, CA 94305, USA. Email: \{alexgruebele, aczerbe, mmcoad, aokamura, cutkosky\}@stanford.edu}%
\thanks{This work was supported in part by National Science Foundation grant 2024247, an ARCS Foundation Fellowship, and the Beijing Institute of Collaborative Innovation.}%
}

\begin{document}

\maketitle
\thispagestyle{empty}
\pagestyle{empty}

\begin{abstract}

Due to their ability to move without sliding relative to their environment, soft growing robots are attractive for deploying distributed sensor networks in confined spaces. Sensing of the state of such robots would add to their capabilities as human-safe, adaptable manipulators. However, incorporation of distributed sensors onto soft growing robots is challenging because it requires an interface between stiff and soft materials, and the sensor network needs to undergo significant strain. In this work, we present a method for adding sensors to soft growing robots that uses flexible printed circuit boards with self-contained units of microcontrollers and sensors encased in a laminate armor that protects them from unsafe curvatures. We demonstrate the ability of this system to relay directional temperature and humidity information in hard-to-access spaces. We also demonstrate and characterize a method for sensing the growing robot shape using inertial measurement units deployed along its length, and develop a mathematical model to predict its accuracy. This work advances the capabilities of soft growing robots, as well as the field of soft robot sensing.

\end{abstract}

\section{Introduction}

Distributed sensor networks are of growing interest for long-term monitoring of environments \cite{yang2002design, mainwaring2002wireless} and structures \cite{balageas2010structural}, but they face limitations in how many sensors can be efficiently deployed, especially in constrained and hard-to-reach spaces. While wireless sensor networks can be designed for low power consumption using batteries \cite{nair2015optimizing} or the ability to scavenge energy \cite{ mathuna2008energy}, they are often unable to transmit data reliably from enclosed spaces such as underground \cite{akyildiz2006wireless} or through walls \cite{damaso2014reliability}.

Soft growing robots \cite{hawkes2017soft} consist of flexible tubes that grow when inflated due to tip eversion, either passively taking on the shape of the environment they grow into, or steered in free space using soft actuators \cite{greer2019soft} or tendons \cite{gan20203d} or with a pre-determined shape \cite{hawkes2017soft}. Due to their flexibility, they can grow to long lengths in highly constrained spaces of unknown shape. To date, sensing for soft growing robots has been primarily focused on the tip of the robot through tools such as cameras \cite{hawkes2017soft,greer2019soft}. Because soft growing robots have a very low cost per unit length and are relatively easy to deploy, they can also be a platform for rapid deployment of sensor networks along the path of the growing robot body.

\begin{figure}[tb]
      \centering
      \includegraphics[width=\columnwidth]{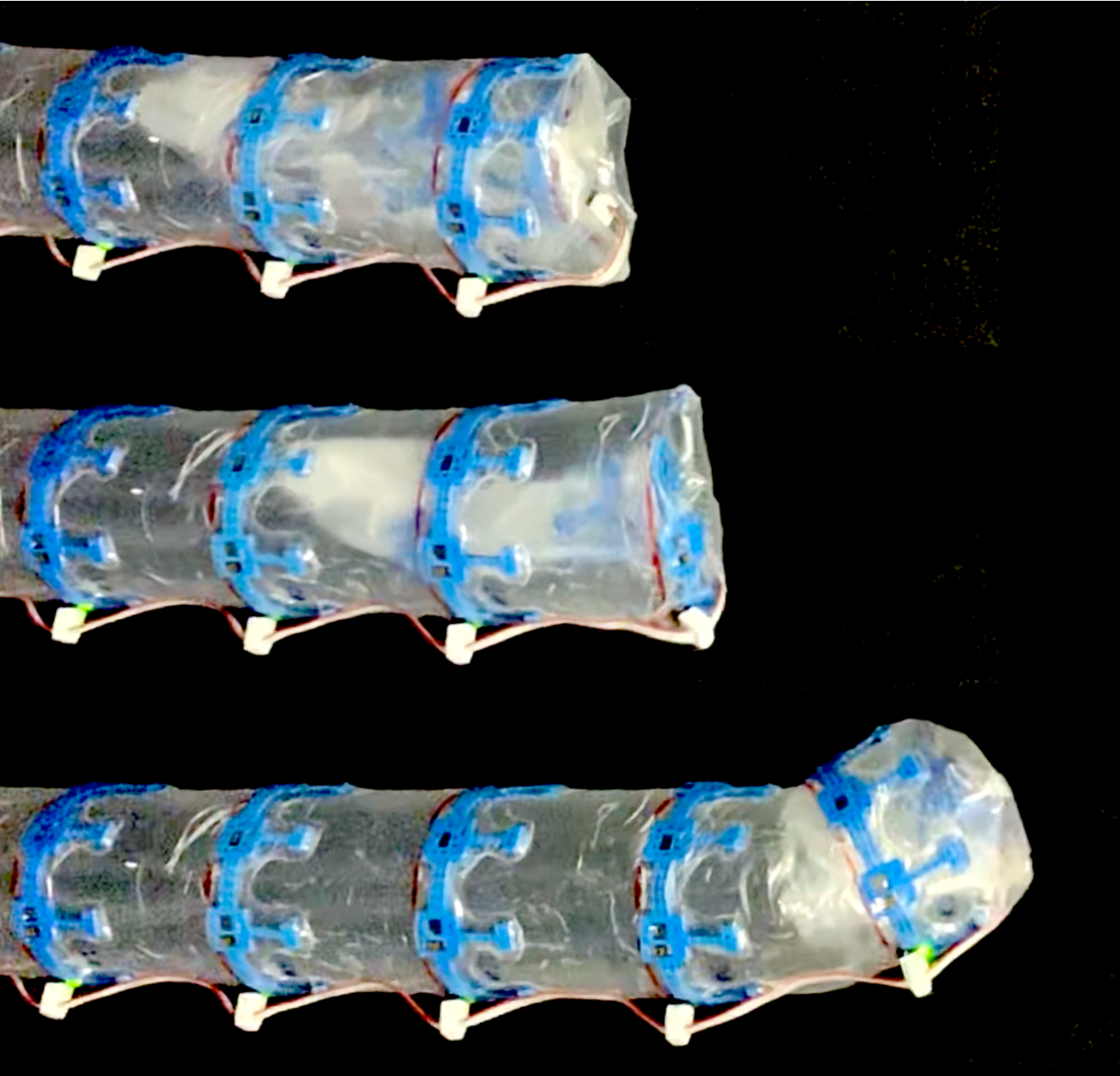}
      \caption{A soft growing robot (semi-transparent) deploying armored sensor bands (blue), and sensing its shape as it traverses a bend.
      }
      \label{glamourshot}
      \vspace{-0.5cm}
  \end{figure}

We use flexible printed circuit board (fPCB) technology to create modular bands of sensors that are distributed along the robot, for continuous monitoring of the immediate environment (Fig.~\ref{glamourshot}). To spatially locate the sources of measurements, the shape of the grown robot is also sensed by measuring the orientation of each sensor band. fPCBs allow for a wide array of traditional MEMS surface-mount sensors to be used without alteration, making this a versatile platform for many types of measurements. In our prototype sensor modules, we include temperature, humidity, acceleration, and orientation. We also demonstrate the capability to determine the spatial direction of a heat source with respect to the robot body.  

Soft growing robots -- and more broadly, many other soft robots -- are inherently unstructured in how they bend and wrinkle. While fPCBs are robust and long-lasting when used within their design parameters, they have shortened lifetimes when subjected to curvatures outside those limits.
Thus, traditional flexible circuits alone are not suitable for placement on soft growing robots. We propose a combination of flexible circuit design and a semi-soft laminate that enforces safe bend radii of the fPCB during uncontrolled wrinkling.
The laminate is designed for high flexibility so as to not hinder robot growth, while protecting the circuit in the directions in which bending occurs. 

\begin{figure*}[!ht]
      \centering
      \includegraphics[width=.9\textwidth]{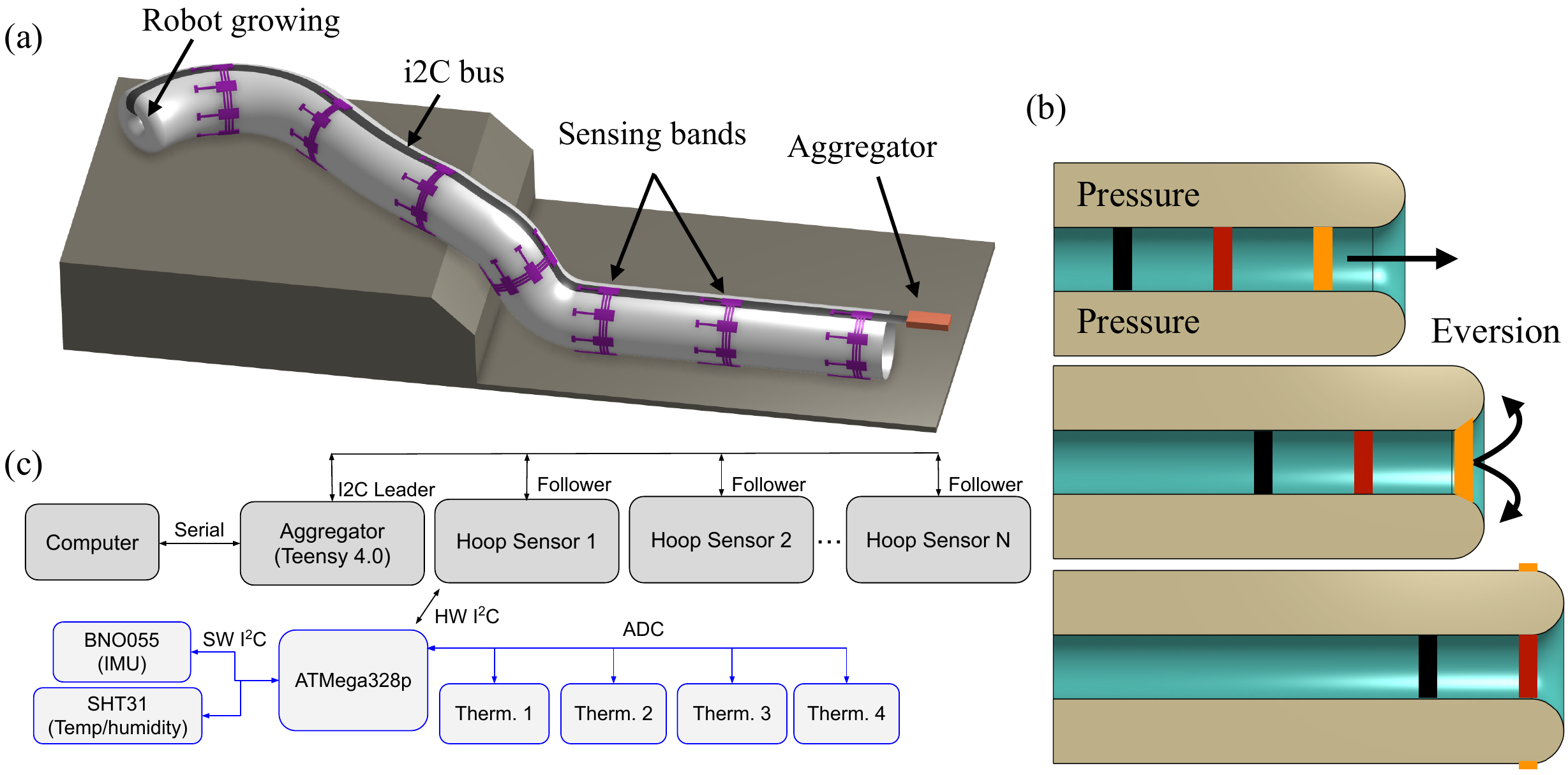}
      \caption{(a) A rendering of a growing robot deploying sensor bands, communicating with an aggregator DAQ via a bus along its axis. (b) The process by which growing robots evert, illustrating the sensor band movement from the stowed state to the deployed state. (c) The overall system architecture, with the components on a single modular band outlined in blue. 
      }
      \label{fig:systemDiagram}
      \end{figure*}
      
In Section \ref{sec:RelatedWork}, we introduce recent advances in scalable sensor network design and soft robot shape sensing. In Section \ref{sec:Design}, we describe the overall system architecture and the laminates used to make the sensors robust. In Section \ref{sec:Experiments}, we characterize the accuracy of the robot’s shape sensing. 
We conclude in Section \ref{sec:Demonstration} with laboratory demonstrations of deploying sensor bands in two applications: (1) locating a leak from a cluster of pipes and (2) exploring a small tunnel.

\section{Related Work}
\label{sec:RelatedWork}

Much of the focus on designing sensor networks has been on outdoor pipeline \cite{stoianov7wireless}, environment \cite{yang2002design}, ecological habitat \cite{mainwaring2002wireless}, and agricultural \cite{liqiang2011crop} monitoring over large areas, necessitating the implementation of wireless data transmission. However, sensor networks are also being deployed in confined spaces that require a higher density of nodes and are close enough to be wired. These operate at a scale where stretchable wiring can be used to expand a network to cover large areas, such as strain gauges for aircraft structural monitoring \cite{chen2018characterization} or on robots \cite{ham2020sensing}. In enclosed spaces, this approach has benefits over wireless transmission, where electromagnetic waves can be attenuated depending on the the obstructing material, such as soil or a wall \cite{akyildiz2006wireless}.

Although the area of coverage of these networks is smaller than outdoor applications, deployment of large numbers of sensor nodes in small areas remains a challenge for both labor efficiency of setup, and precise knowledge of where the nodes (and therefore data source) are located spatially. One approach is to make the sensor nodes so small that they are simply scattered and unobtrusive \cite{warneke2002autonomous, scott2003ultralow}, though these still face the challenge of localizing the data source. By introducing inertial measurement units (IMUs) to sensor nodes on a flexible tape, \cite{dementyev2015sensortape} were able to localize sensor readings by determining shape of the carrier tape in real time, in a similar manner to what we present. By including the entire network on a single continuous strip, the distance that it could span was limited and it was not designed for robotic deployment.

Soft growing robots have been used for sensor deployment including cameras \cite{hawkes2017soft} and antennas \cite{blumenschein2018tip}, however work has been primarily focused on adding sensors to a tip mount \cite{jeong2019tip, CoadRAM2020}. While this approach is useful for taking data along the path of the robot tip as it grows into constrained spaces, it does not use the length of robot left behind after growth for long-term monitoring of the space it grew into. At a shorter scale, \cite{gonzalez2010toward} used capacitive touch sensors distributed along the length of a robot.

In many scenarios in which soft growing robots are used to deploy sensors, the shape of the robot must be measured to localize each node. Approaches to measuring curvature of soft robots include optical \cite{searle2013optical}, capacitive \cite{shahmiri2020sharc}, resistive \cite{truby2020distributed, tapia2020makesense}, and mechanical methods using cables on encoders \cite{zou2020design}. All of these approaches require substantial sensor integration in the robot's fabrication process, and are challenging to integrate over the length scales reachable with soft growing robots.

\section{Sensor Design}
\label{sec:Design}

Building on and addressing gaps in related work, we present a distributed sensor network that can be deployed using soft growing robots.

\subsection{Design objectives}

There are two purposes for incorporation of sensors onto robots: exteroception (that is, sensing of external stimuli), and proprioception (that is, sensing of the state of the robot itself). We sought to enable both of these purposes for soft growing robot sensing with a single design. For exteroception, a suite of sensors could be useful to allow monitoring of confined spaces. Depending on the application, sensing of temperature, humidity, light, and video could all be useful. In addition, by sensing the shape and contact conditions of the robot itself when deployed in an environment, information can be learned about the environment's shape and stiffness properties. For proprioception, sensing of robot shape and contact conditions can improve control of the robot during navigation and manipulation tasks. For some scenarios, sensors along one side of the robot body would suffice, but for other scenarios, sensors should be placed on all sides of the robot body to achieve the desired functionality.

We aimed to create a template sensor system in which the specific sensors could be switched out as needed, depending on the application. To demonstrate proprioception, we chose to place IMUs along the length of the robot to sense its shape. To demonstrate exteroception and directional sensing capability, we placed temperature and humidity sensors along the length of the robot and thermistors distributed on all sides of the robot body. In future iterations of the design, these specific sensors could be replaced with cameras and contact sensors, for example, as needed.

In order to function on a soft growing robot, our sensor system needed to fulfill the following basic requirements. First, it needed to be attached to the soft robot body and endure repeated eversion and inversion without failure or significant encumbrance of the robot's ability to grow and retract. Second, it needed to be scalable to robot lengths suitable for navigation and manipulation tasks (ranging from 1 to 10 meters in length). 

\subsection{System overview}

The sensor network (Fig. \ref{fig:systemDiagram}) consists of an aggregator microcontroller (Teensy 4.0) located at the base of the robot and many fPCB sensor bands distributed along its length at discrete intervals. The sensor bands are adhered to the outside of the robot so that the sensors can monitor their direct surroundings when deployed. Each band is self-contained with its own microcontroller (Atmega328P-MU), reducing noise by performing digital conversions over short distances from analog sensors. Communication to the aggregator microcontroller is performed through a 4-wire i2C bus which runs the full length of the robot. This system is robust to failure of individual sensor bands, and uses a small number of flexible wires.

\begin{figure}[!t]
      \centering
      \includegraphics[width=\columnwidth]{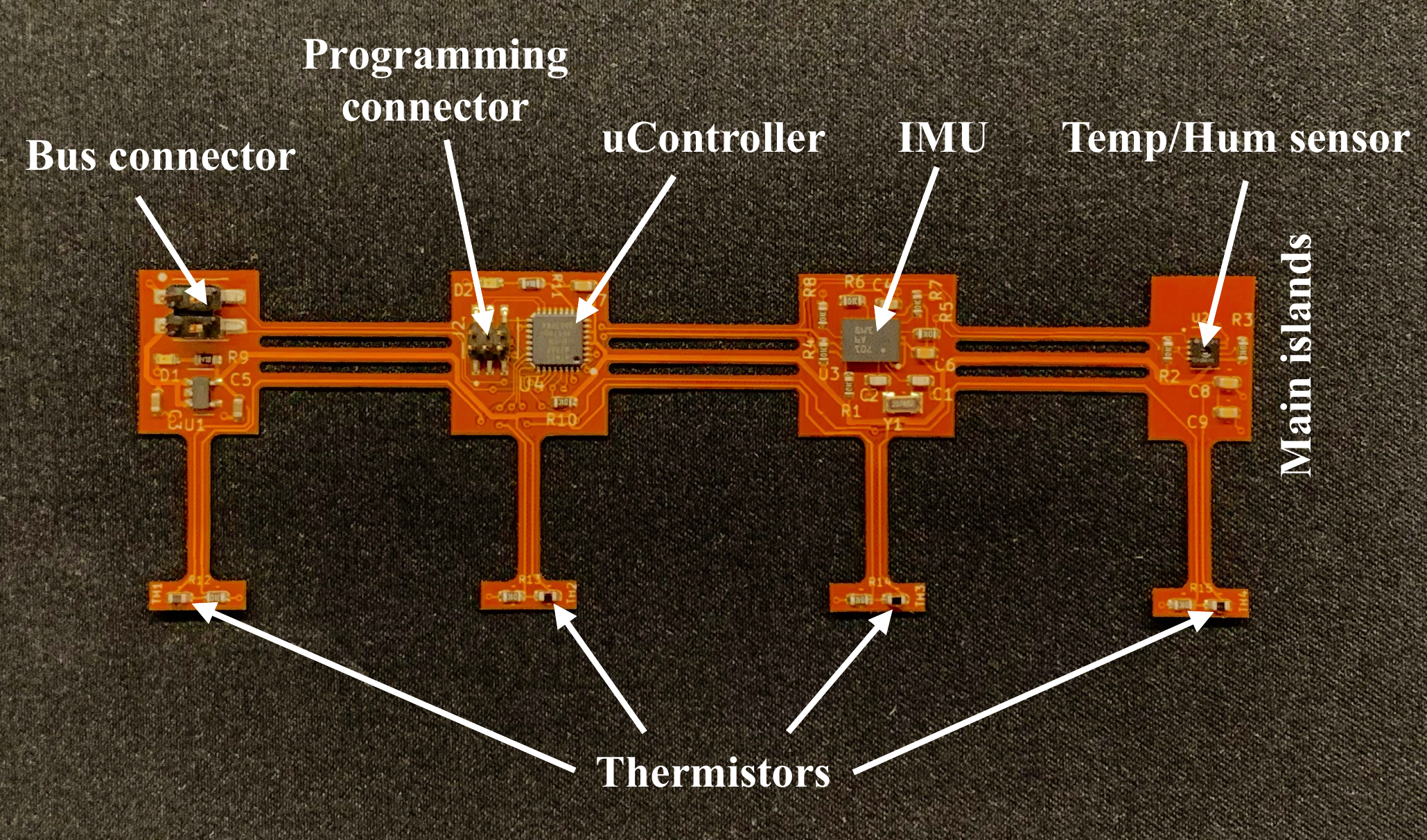}
      \caption{A flexible PCB band designed for growing robot deployment. Components are distributed in multiple semi-rigid islands, connected by highly flexible tracks with a sparsely filled ground layer. Many components that require directional sensing (such as the thermistors here) can be placed along additional islands which align with the axis of the robot.}
      \label{fig:sensorBand}
  \end{figure}

The sensor bands (Fig. \ref{fig:sensorBand}) consist of semi-rigid islands that contain the surface mount components, and flexible tracks which carry circuit traces between islands (fPCBs were produced by Seeed Studio with a 0.1~mm thick polyimide backing and cross-hatch ground pour beneath the flexible tracks for minimally stiff shielding). This design is chosen so that the bands can be wrapped circumferentially around the robot while exhibiting high flexibility so as not to impede growth of the robot or damage the circuit. The flexible islands are in a two-dimensional network: the main islands consist of the non-directional components (I2C devices, microcontroller, peripheral components, and bus-connector); the smaller islands contain the directional sensors. This approach is scalable, as it allows for a chain of directional sensors to be added down the axial direction of the robot, rather than increasing the size of the rigid islands (and therefore reducing band flexibility). The bands are attached to the robot with Tegaderm adhesive film, resulting in a robust adhesion that still allows the flexible tracks to bend during eversion. \alex{This highly conformable and thin adhesive allows the bands to be fully covered so that they do not snag on the environment, and does not rely on connection directly to the polyurethane armor, a material which is inherently difficult to adhere to.}

Sensors on each band include an IMU (Bosch BNO055), a temperature/humidity sensor (Sensirion SHT31-DIS-B2.5kS), and four thermistors (TDK NTCG163JF103FT1S). In other applications, measurements that could be simply integrated into our platform using standard SMD chips include contact, force, air pressure, ambient light, gas composition, pH, and proximity. 


\subsection{PCB armor}

While flexible circuit technology has been ubiquitous in industry use since the 1950s, it is only recently finding use cases in soft robotics \cite{wallin20203d}. Since soft robots often undergo unpredictable motions, flexible sensors adhered to these robots are at risk of undergoing bends and wrinkles that would break fragile connections to surface mount components and the conductive traces that connect them. Most manufacturers recommend that dynamic flex circuits have a minimum bend radius over traces of at least 100 times board thickness so that the copper does not work harden or succumb to cyclic fatigue \cite{pcbdesign}.

\begin{figure}[!t]
      \centering
      \includegraphics[width=\columnwidth]{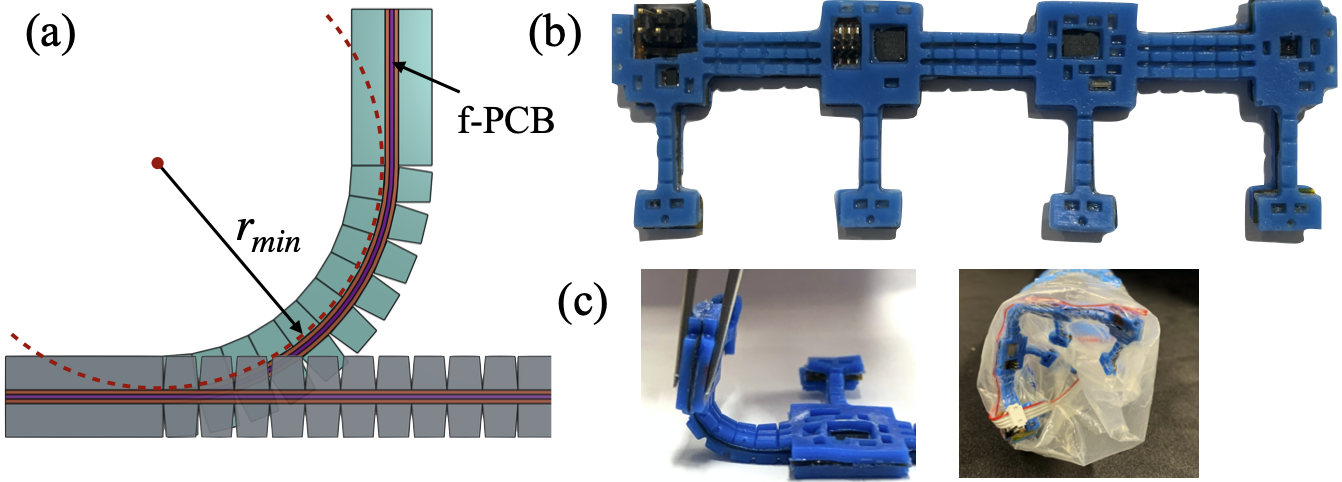}
      \caption{The elastomer armor enforces a minimum bend radius while maintaining high flexibility, only jamming at the necessary curvature. The armor is molded as one piece per side, and laminated onto the fPCB using an adhesive-backed Kapton intermediate layer. \alex{For this design, the angle between adjacent repeating teeth when laid flat is $19^{\circ}$.}
      }  
      \label{armor}
  \end{figure}

In order to enforce this minimum bend radius, we designed an armor laminate to protect the fPCBs (Fig. \ref{armor}), which encases the circuits on top and bottom. Over the functional islands, the laminate is stiff and has cavities that fit snugly over the SMD components, protecting them from damage and providing strain relief to the soldered connections. Over the flexible tracks, the laminate has very low stiffness (so as to not hinder deployment) until the minimum bend radius is reached, at which point the teeth mesh and prevent further bending (Fig. \ref{armor} (a)). Tooth geometry can be varied depending on the allowable bend in the application. The laminate is fabricated by pouring Vytaflex 60 \alex{(60A Shore hardness \cite{smoothon})} polyurethane into a 3D printed mold. Once cured, it is adhered to a 3 mil thickness Kapton sheet with adhesive backing, and this is then applied to the top and bottom of the sensor band.


\subsection{Shape accuracy and model}

\margaret{To estimate the robot's shape based on the measured IMU orientation at each sensor band, some assumptions are needed about the shape of the robot body segments between the sensor bands.} 

\margaret{First, because the robot body material is nearly inextensible, we assume that the outer edge of the robot body remains constant in length as the robot curves, while the inner edge wrinkles. This is consistent with the robot's observed behavior when pressurized, and thus is expected to be the case when the robot is growing.} 

\begin{figure}[!t]
      \centering
      \includegraphics[width=\columnwidth]{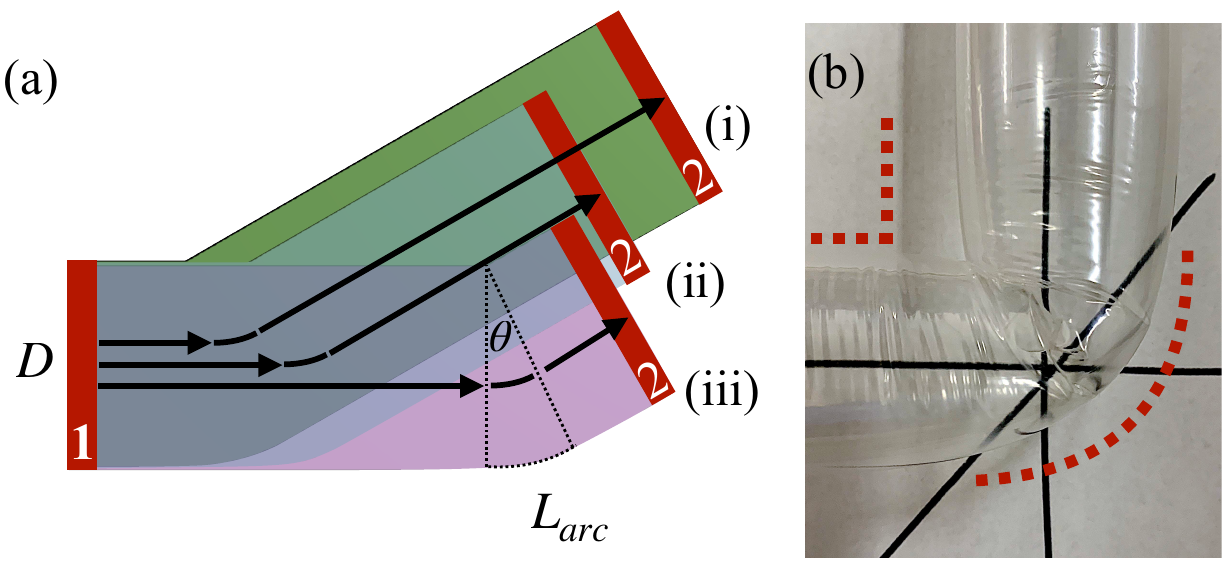}
      \caption{(a) The robot is assumed to bend a maximum of once between sensor bands (red). The position between the sensors at which it bends is unknown and can contribute to error in the robot's shape measurement, varying between a bend immediately after the first sensor (i) to right before the following sensor (iii). (b) The robot forms smooth outer bends when pressurized for angles up to 90$^{\circ}$.
      }
      \label{fig:model_schematic}
  \end{figure}

\margaret{Second, we assume that a single discrete bend takes place between each pair of sensors, with all of the length change of the inner edge of the robot body taken up at a single point (Fig.~\ref{fig:model_schematic} (a)). This strain profile is consistent with inflated beam bending behavior, and a single bend in each segment is likely provided that the sensor bands are spaced close enough for the particular operating pressure and robot diameter used. In the case of the robot used here, a ratio of band spacing to diameter of 1.5 was found to prevent multiple bends per segment.}

\margaret{Third, to account for the amount of length taken up by the curved sections of the robot, we assume that the outer edge of the robot body forms a circular arc at each bend, with straight segments taking up the rest of the length between sensor bands. We found this to be consistent with the robot's observed behavior when pressurized for angles up to about 90$^{\circ}$ (Fig.~\ref{fig:model_schematic} (b)).}

\margaret{And fourth, we assume that the orientation of the central axis of the robot body can be accurately measured by the orientation of any of the corresponding points on its surface, i.e., that it does not matter where around the circumference of the robot body the IMU is placed. This assumption is consistent with robot behavior provided that the environment around the robot does not cause a depression in the robot body that happens to be where the IMU is located. To avoid this, we ensured that, for our experiments, the robot body was smaller in diameter than the smallest aperture in the environment it would explore.} 

\margaret{Given these assumptions, shape sensing inaccuracy can come from: (1) uncertainty in where bends occur between sensor bands, and (2) inaccuracy in measurements from the IMUs themselves. Further inaccuracy will come into play if the previous assumptions are not true.}

\margaret{For a large enough band spacing, there is inherent uncertainty in the position of each sensor band relative to the previous one, because the location of each bend within the segment between successive sensor bands is unknown. The bend could occur immediately after the first sensor, as shown in Fig.~\ref{fig:model_schematic} (a) (i), right before the next sensor (iii), or somewhere in between (ii). Because bends are equally likely at any point along this length, the safest assumption by the central limit theorem is that bends occur at mean. Thus we choose the midpoint of the segment as the center of each bend for our shape reconstruction algorithm.} To quantify the shape error due to  uncertainty in the bend location, we use a model that is mostly influenced by band spacing, $L_{spacing}$, but also a function of robot diameter, $D$ (Fig.~\ref{fig:model_schematic} (a)). The length over which the bend can begin is limited by the severity of the bend angle and the diameter of the robot as $L_{arc} = (D/2) \theta$, where  $\theta$ is the bend angle. Thus, the range over which the bend can occur is $L_{spacing}-L_{arc}$.

Another factor that influences measured shape accuracy is IMU accuracy. While accelerometers can be used to very accurately measure IMU angle relative to gravitational acceleration, the calculation of IMU heading \andrew{still uses the chip's sensor fusion algorithm but primarily} relies on a magnetometer, which is comparatively less accurate; the BNO055 sensor used has a magnetometer heading accuracy of $\pm$2.5$^\circ$. In addition, the magnetometer can require re-calibration when the magnetic fields in its environment change, often with the introduction of a ferrous or magnetic object, which will result in a significant change in reported heading.

\margaret{Based on these considerations,} the choice of band spacing is driven either by the required spatial resolution of the measurements or by the desired accuracy of the sensed robot shape, at the cost of sensor cost and measurement frequency.

\section{Experiments and Results}
\label{sec:Experiments}

  \begin{figure*}[!htb]
      \centering
      \includegraphics[width=\textwidth]{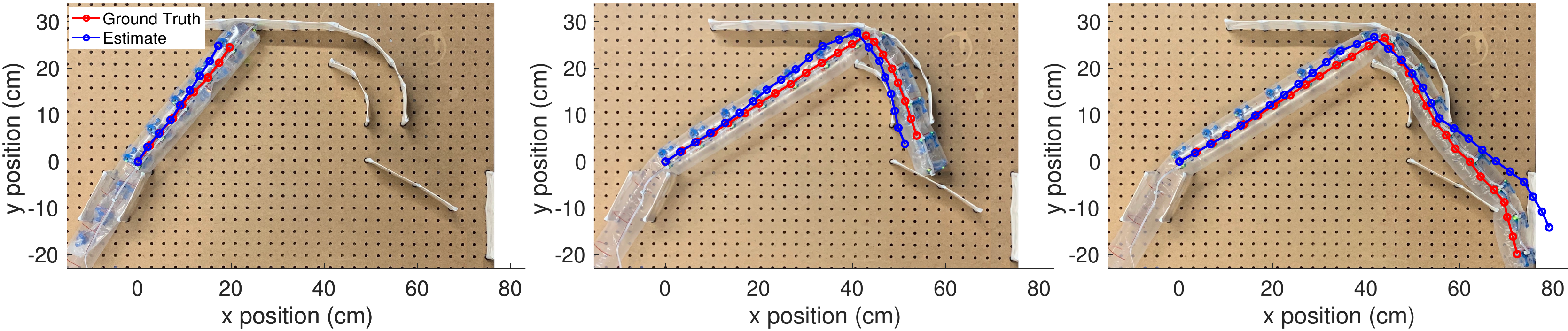}
      \caption{Plots of ground truth (from photo) and estimated robot shape at three time points during maze navigation. At each step, the maximum position error for any point on the robot body is less than 9~cm, with a total of 76~cm of robot deployed. \alex{Note that the shape of the robot's deployed section changes due to external forces -- thus sensing orientation with a sensor just at the moving tip would not provide live robot shape in open space like this.}
      }
      \label{fig:maze_growth}
  \end{figure*}
  
\begin{figure}[!htb]
      \centering
      \includegraphics[width=0.66\columnwidth]{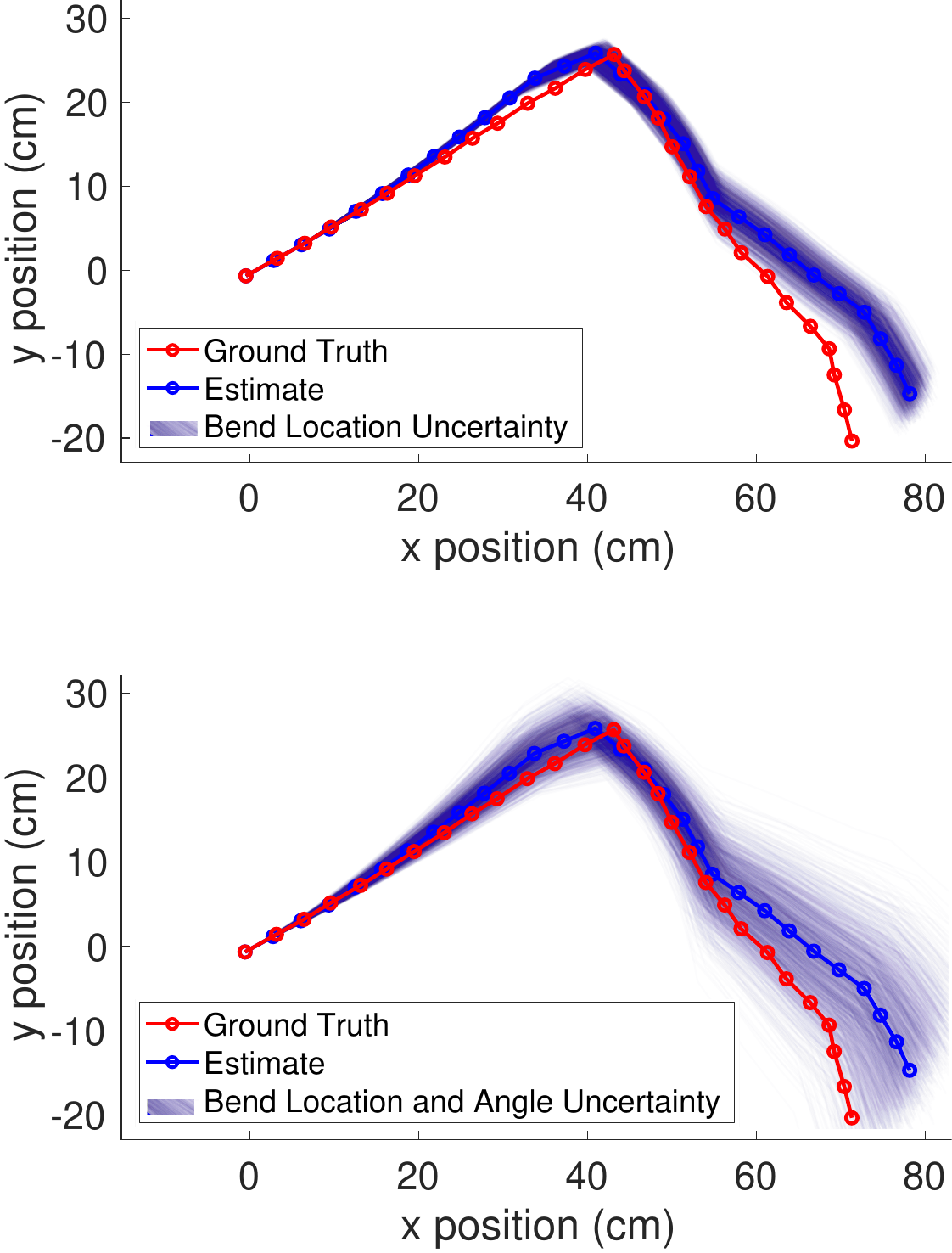}
      \caption{Plots showing uncertainty in the sensed shape from third time point in Fig.~\ref{fig:maze_growth} based on two sources: (top) uncertainty in the bend location, and (bottom) uncertainty in the bend location as well as uncertainty in the sensed bend angle ($\pm 3^{\circ}$). The deviation of our shape estimate from the ground truth shape comes primarily from the uncertainty in the sensed angle, which could be improved with more accurate or redundant sensors.
      }
      \label{fig:maze_uncertainty}
      \vspace{-0.5cm}
  \end{figure}

To validate sensor network deployment and shape sensing accuracy, we placed 15 sensor bands at a 7.6~cm (3~in) spacing along a 1.15~m robot of diameter 6.6~cm. The bands were connected via a bus to an aggregator microcontroller, which sent measurements at 50Hz to a computer running custom real-time plotting and recording software using Python. We built a maze on a peg-board through which the robot was able to grow. While we tested the robot's growth capabilities separately, to ensure that the sensors would survive this test, we posed the robot in three positions that it would reach during growth through the maze and captured the sensor data in those positions, rather than during an active growth. The maze forces the robot to form multiple bends of different angles, as well as to follow a trajectory such that the deployed section of the robot changes shape as it continues to grow. In a scenario like this, sensors moving with the tip of the robot would not be able to measure shape changes further back, so distributed sensing is helpful to accurately capture the robot shape. 

To measure the ground truth shape, we placed a camera above the maze and took photos that could then be fed into MATLAB, where the robot shape was reconstructed by clicking on key points in the image. After clicking on two points to calibrate the pixel-to-world distance measurement and define the angle of rotation of the maze relative to the image frame, we traced out the robot shape by clicking on one point at the center of each sensor band, as well as one point along the robot's center line between each pair of sensor bands. 

We calibrated the IMUs with their z-axes pointing upwards so that it was easy to project the measured shape into the plane of the maze by projecting it into the x-y plane. We zeroed all IMUs while pointing in the same direction at the start to avoid any variability between sensors. We calculated the estimated robot shape by feeding the measured IMU orientations into the model discussed in Section~\ref{sec:Design}. We then projected the sensed 3D shape into the plane of the maze. To register the estimated robot shape to the image, we rotated the estimated shape \margaret{within the plane of the maze} until the vector from the center of the first to the center of the second sensor band was aligned with the same vector in the ground truth plot.
\margaret{While soft growing robots tend not to twist significantly about their axis, small twist angles can build up as the robot gets longer. Our shape sensing method still works if portions of the robot twist relative to the base, and the IMU readings could be used to measure this twist.} 

Fig.~\ref{fig:maze_growth} shows the images from the three chosen time points of maze navigation, with the ground truth shape overlaid in red, and the estimated shape overlaid in blue. In the first image, the maximum position error between corresponding ground truth and estimated shape points is 2.5~cm and occurs at the sensor band closest to the robot tip. In the second image, the maximum position error is 3.6~cm and occurs midway along the grown length of the robot. In the third image, the maximum position error is 8.93~cm and occurs at the sensor band closest to the tip.

To understand the sources of error between the ground truth shape and our estimated shape, we used our model to conduct Monte Carlo simulations of the estimated shape when two parameters were varied: the bend location, and the sensor-measured orientation. The first parameter, bend location, is inherently uncertain, even if the sensors are perfectly accurate in their orientation readings, because the bend location could be anywhere between each pair of sensors. To vary the bend location, we calculated 2000 robot shapes with the exact bend angles measured by the IMUs but with bend locations sampled from a uniform distribution along the length between each pair of sensors \margaret{minus the length taken up by the bend itself}. The results of this simulation for the third time point are plotted in Fig.~\ref{fig:maze_uncertainty} (top). The simulated shapes are plotted in translucent blue such that the intensity of the color corresponds to the likelihood of the shape. Even across the entire deployed length of the robot, the expected position error due to this factor is quite small and does not fully explain the measured error.

The second parameter, sensor-measured orientation, is dependent on the quality of the sensors used. To vary this parameter in addition to the bend location, we calculated 2000 robot shapes with the sensor-measured bend angles plus a sensor error sampled from a uniform distribution between $\pm$3$^\circ$, as well as with bend locations sampled from a uniform distribution along the length between each pair of sensors \margaret{minus the length taken up by the bend itself}. The results of this simulation for the third time point are plotted in Fig.~\ref{fig:maze_uncertainty} (bottom). This amount of sensor error plus the uncertainty of bend location plausibly explains the error in our estimated robot shape. Such a simulation technique can be used to design the sensor band spacing and sensor error tolerance for a growing robot system for shape sensing with a particular position error tolerance.

Deployable sensor networks are of interest in many long term monitoring applications. We demonstrate two applications in which the robot can provide data from places that are otherwise hard to access.

  \begin{figure*}[!ht]
      \centering
      \includegraphics[width=\textwidth]{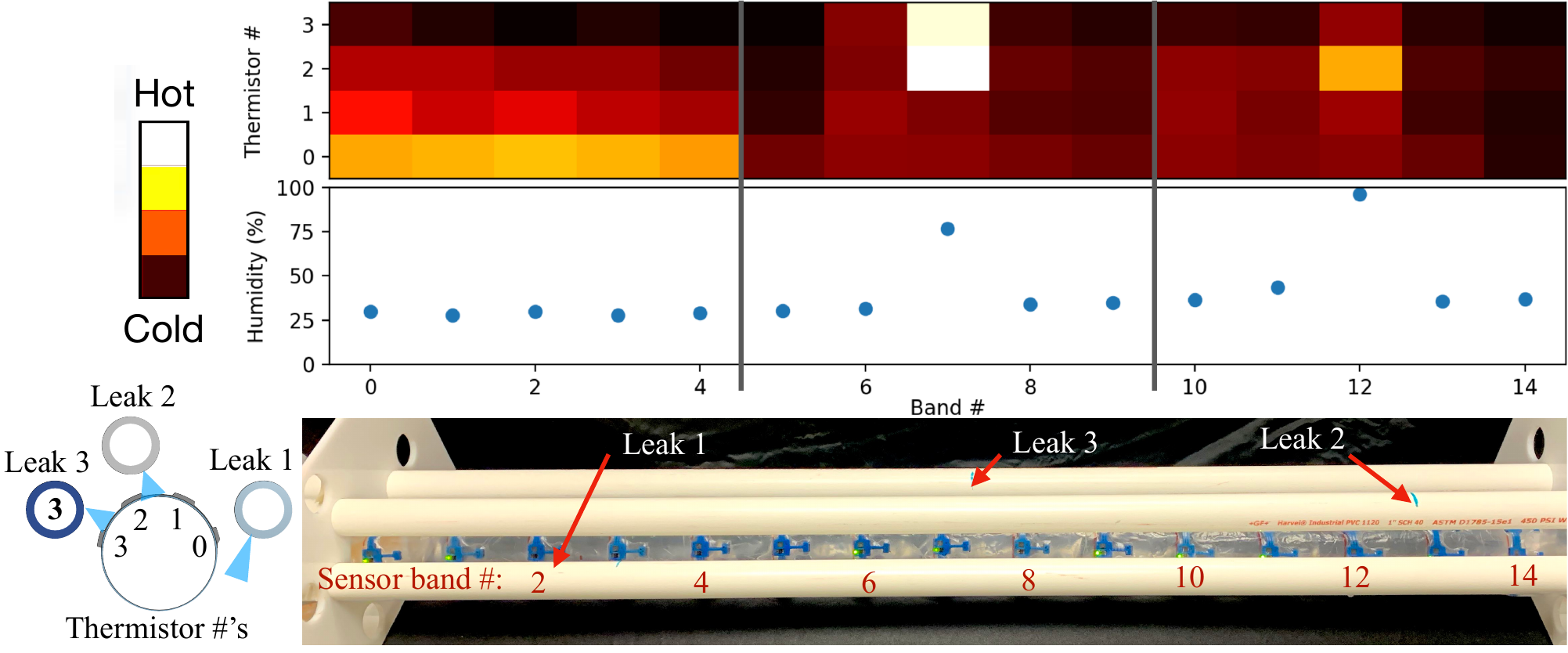}
      \caption{The robot is deployed between three steam pipes, such that the thermistor islands face the pipes as shown. A steam leak is triggered in each of the three pipes, in an orientation to the robot as shown (inset). Humidity and heat direction are recorded at nearby sensors and plotted, with vertical lines demarcating the relevant area for each of the three separate trials.
      }
      \label{fig:steam}
  \end{figure*}

\section{Demonstration}
\label{sec:Demonstration}

\subsection{Deployable Temperature and Humidity Sensing}

In our first demonstration, we deploy the robot between a tight cluster of steam pipes to identify the location of leaks. Using a small entry point to enter the wall and pinpoint the leak location can reduce the amount of damage done to the wall in order to access the pipes behind it. The robot is able to grow in the constrained space between the pipes, and measure humidity and temperature over time, with thermistors measuring temperature at four discrete intervals encompassing the top half of the robot's circumference. \margaret{Note that the melting temperature of LDPE plastic is at least $105^\circ$C, so the robot body can withstand the temperature of steam at atmospheric pressure without melting \cite{polymerdatabase}.} The setup and measured response are shown in Fig.~\ref{fig:steam}. The three steam pipes are oriented as shown in the inset, with vectors indicating the leak holes; the fifteen sensor bands are numbered along the length, and the four thermistors per band are pointed as indicated in the inset, with the humidity sensor located near thermistor 3. Steam is released into one pipe at a time for three separate trials, and the neighboring 5 sensors' humidity and thermistor values are plotted for each. 

The first pipe is oriented such that the steam leak is nearest to sensor band 2, but entirely blocked from the humidity sensor, which shows no noticeable rise in humidity. The steam is also pointed away from the robot entirely, resulting in a measurable but low thermistor change, and heat from the steam and pipe percolates to nearby bands. Thermistor 0 of band 2 is still identifiable as a local maximum in temperature, so with the robot's orientation, the leak can be identified. Leaks 2 and 3 are pointed more directly at the robot, and thus register a much higher peak in thermistor value, which can be clearly identified to register leak location at bands 7 and 12. The orientations of the heat sources as registered to the thermistor placement show that the leaks occur in pipes 2 and 3 respectively. 

\begin{figure}[!ht]
      \centering
      \includegraphics[width=\columnwidth]{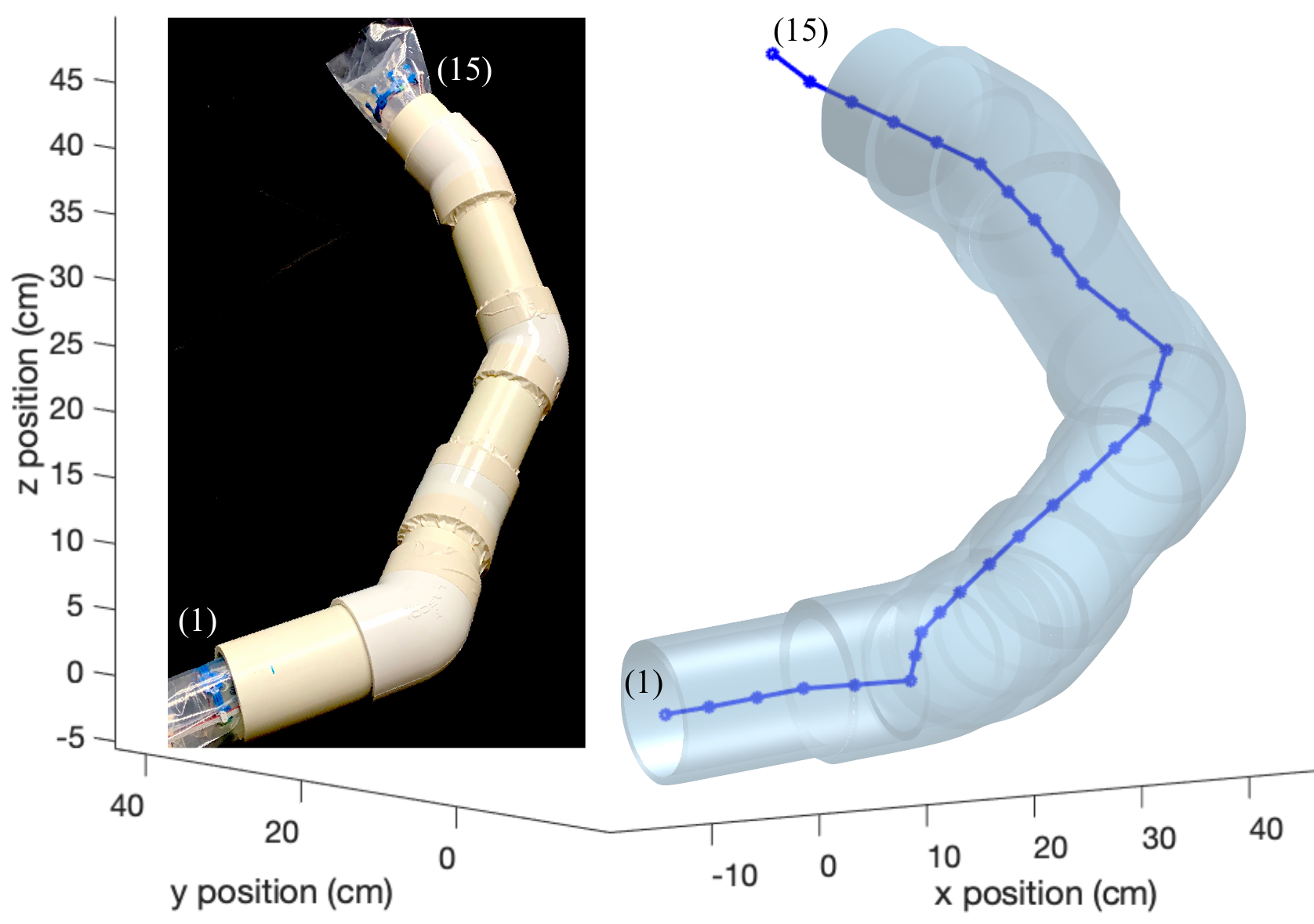}
      \caption{(left) The 6.6 cm diameter robot inside a 100 cm long PVC pipe of 7.6 cm internal diameter (Size 3). (right) A CAD model of the same pipe, with the shape estimated by 15 sensors overlaid in blue.}
      \label{fig:Shape}
  \end{figure}

\subsection{Shape sensing in a constrained space}

In our second demonstration we show shape sensing in three dimensional space by \margaret{placing the robot in} a constrained pipe such that the first sensor band is at the entry point, and the last comes out the exit (Fig. \ref{fig:Shape}). This demonstration mimics growth into an unknown path such as an animal burrow, for mapping tunnels in ecological studies. The pipe has an internal diameter of 7.6 cm, and traverses two $22.5^\circ$ bends and two $45^\circ$ bends, which are not co-planar. The shape calculated using our model and based on the 15 sensor orientations stays within the 7.6 cm diameter throughout its length.



\section{Conclusion and Future Work}
We created and demonstrated a distributed fPCB sensor network deployed on a soft growing robot. IMU chips on each band allow for shape reconstruction of the soft robot using absolute orientation, and a simple model is developed to determine and reduce shape uncertainty. A network of digital temperature and humidity sensors as well as thermistors allows for humidity sensing along the robot and directional temperature information to localize heat sources in space. These sensors are protected with an armor laminate that jams at a minimum bend radius to protect the fPCB during undeployed states, and robot tip eversion. This approach is easily scalable, and its modularity allows for the usage of many different types of sensors, depending on the sensing application.
This sensor network opens up new possibilities for control of soft growing robots. In combination with soft actuators such as series pouch motors \cite{CoadRAM2020}, 
the robot could be steered to a specific shape or endpoint position. With the addition of directional tactile sensing (for example in place of the thermistors), contact sensing could be used to navigate an environment. Shape sensing itself could be improved significantly with the addition of a redundant IMU on each band for increased accuracy, as well as a separate type of band that only includes the IMU(s), distributed at a higher density. \andrew{These large arrays could also be improved with a selectively stiff armor design, since fully rigid islands would protect solder connections from cracking.} In general, the development of multiple types of bands distributed at different densities, depending on the sensing type, would allow for greater capability without needlessly increased cost and complexity. Finally, the implementation of a longer-range protocol such as RS485 over the main bus would allow for arbitrarily long robot lengths with negligible performance impacts. 






\bibliographystyle{IEEEtran}
\bibliography{references}

\begin{thebibliography}{10}
\providecommand{\url}[1]{#1}
\csname url@samestyle\endcsname
\providecommand{\newblock}{\relax}
\providecommand{\bibinfo}[2]{#2}
\providecommand{\BIBentrySTDinterwordspacing}{\spaceskip=0pt\relax}
\providecommand{\BIBentryALTinterwordstretchfactor}{4}
\providecommand{\BIBentryALTinterwordspacing}{\spaceskip=\fontdimen2\font plus
\BIBentryALTinterwordstretchfactor\fontdimen3\font minus
  \fontdimen4\font\relax}
\providecommand{\BIBforeignlanguage}[2]{{%
\expandafter\ifx\csname l@#1\endcsname\relax
\typeout{** WARNING: IEEEtran.bst: No hyphenation pattern has been}%
\typeout{** loaded for the language `#1'. Using the pattern for}%
\typeout{** the default language instead.}%
\else
\language=\csname l@#1\endcsname
\fi
#2}}
\providecommand{\BIBdecl}{\relax}
\BIBdecl

\bibitem{yang2002design}
X.~Yang, K.~G. Ong, W.~R. Dreschel, K.~Zeng, C.~S. Mungle, and C.~A. Grimes,
  ``Design of a wireless sensor network for long-term, in-situ monitoring of an
  aqueous environment,'' \emph{Sensors}, vol.~2, no.~11, pp. 455--472, 2002.

\bibitem{mainwaring2002wireless}
A.~Mainwaring, D.~Culler, J.~Polastre, R.~Szewczyk, and J.~Anderson, ``Wireless
  sensor networks for habitat monitoring,'' in \emph{Proceedings of the 1st ACM
  international workshop on Wireless sensor networks and applications}, 2002,
  pp. 88--97.

\bibitem{balageas2010structural}
D.~Balageas, C.-P. Fritzen, and A.~G{\"u}emes, \emph{Structural health
  monitoring}.\hskip 1em plus 0.5em minus 0.4em\relax John Wiley \& Sons, 2010,
  vol.~90.

\bibitem{nair2015optimizing}
K.~Nair, J.~Kulkarni, M.~Warde, Z.~Dave, V.~Rawalgaonkar, G.~Gore, and
  J.~Joshi, ``Optimizing power consumption in iot based wireless sensor
  networks using bluetooth low energy,'' in \emph{2015 International Conference
  on Green Computing and Internet of Things (ICGCIoT)}.\hskip 1em plus 0.5em
  minus 0.4em\relax IEEE, 2015, pp. 589--593.

\bibitem{mathuna2008energy}
C.~O. Mathuna, T.~O’Donnell, R.~V. Martinez-Catala, J.~Rohan, and
  B.~O’Flynn, ``Energy scavenging for long-term deployable wireless sensor
  networks,'' \emph{Talanta}, vol.~75, no.~3, pp. 613--623, 2008.

\bibitem{akyildiz2006wireless}
I.~F. Akyildiz and E.~P. Stuntebeck, ``Wireless underground sensor networks:
  Research challenges,'' \emph{Ad Hoc Networks}, vol.~4, no.~6, pp. 669--686,
  2006.

\bibitem{damaso2014reliability}
A.~D{\^a}maso, N.~Rosa, and P.~Maciel, ``Reliability of wireless sensor
  networks,'' \emph{Sensors}, vol.~14, no.~9, pp. 15\,760--15\,785, 2014.

\bibitem{hawkes2017soft}
E.~W. Hawkes, L.~H. Blumenschein, J.~D. Greer, and A.~M. Okamura, ``A soft
  robot that navigates its environment through growth,'' \emph{Science
  Robotics}, vol.~2, no.~8, 2017.

\bibitem{greer2019soft}
J.~D. Greer, T.~K. Morimoto, A.~M. Okamura, and E.~W. Hawkes, ``A soft,
  steerable continuum robot that grows via tip extension,'' \emph{Soft
  robotics}, vol.~6, no.~1, pp. 95--108, 2019.

\bibitem{gan20203d}
L.~T. Gan, L.~H. Blumenschein, Z.~Huang, A.~M. Okamura, E.~W. Hawkes, and J.~A.
  Fan, ``3d electromagnetic reconfiguration enabled by soft continuum robots,''
  \emph{IEEE Robotics and Automation Letters}, vol.~5, no.~2, pp. 1704--1711,
  2020.

\bibitem{stoianov7wireless}
I.~Stoianov, L.~Nachman, and S.~Madden, ``A wireless sensor network for
  pipeline monitoring,'' in \emph{IPSN}, vol.~7, 2007, pp. 25--27.

\bibitem{liqiang2011crop}
Z.~Liqiang, Y.~Shouyi, L.~Leibo, Z.~Zhen, and W.~Shaojun, ``A crop monitoring
  system based on wireless sensor network,'' \emph{Procedia Environmental
  Sciences}, vol.~11, pp. 558--565, 2011.

\bibitem{chen2018characterization}
X.~Chen, T.~Topac, W.~Smith, P.~Ladpli, C.~Liu, and F.-K. Chang,
  ``Characterization of distributed microfabricated strain gauges on
  stretchable sensor networks for structural applications,'' \emph{Sensors},
  vol.~18, no.~10, p. 3260, 2018.

\bibitem{ham2020sensing}
J.~Ham, ``Skin-like multi-modal sensing devices for dexterous robotic hands,''
  Ph.D. dissertation, Stanford University, 2020.

\bibitem{warneke2002autonomous}
B.~A. Warneke, M.~D. Scott, B.~S. Leibowitz, L.~Zhou, C.~L. Bellew, J.~A.
  Chediak, J.~M. Kahn, B.~E. Boser, and K.~S. Pister, ``An autonomous 16 mm/sup
  3/solar-powered node for distributed wireless sensor networks,'' in
  \emph{SENSORS, 2002 IEEE}, vol.~2.\hskip 1em plus 0.5em minus 0.4em\relax
  IEEE, 2002, pp. 1510--1515.

\bibitem{scott2003ultralow}
M.~D. Scott, B.~E. Boser, and K.~S. Pister, ``An ultralow-energy adc for smart
  dust,'' \emph{IEEE Journal of Solid-State Circuits}, vol.~38, no.~7, pp.
  1123--1129, 2003.

\bibitem{dementyev2015sensortape}
A.~Dementyev, H.-L. Kao, and J.~A. Paradiso, ``Sensortape: Modular and
  programmable 3d-aware dense sensor network on a tape,'' in \emph{Proceedings
  of the 28th Annual ACM Symposium on User Interface Software \& Technology},
  2015, pp. 649--658.

\bibitem{blumenschein2018tip}
L.~H. Blumenschein, L.~T. Gan, J.~A. Fan, A.~M. Okamura, and E.~W. Hawkes, ``A
  tip-extending soft robot enables reconfigurable and deployable antennas,''
  \emph{IEEE Robotics and Automation Letters}, vol.~3, no.~2, pp. 949--956,
  2018.

\bibitem{jeong2019tip}
S.-G. Jeong, M.~M. Coad, L.~H. Blumenschein, M.~Luo, U.~Mehmood, J.~H. Kim,
  A.~M. Okamura, and J.-H. Ryu, ``A tip mount for carrying payloads using soft
  growing robots,'' \emph{arXiv preprint arXiv:1912.08297}, 2019.

\bibitem{CoadRAM2020}
M.~M. Coad, L.~H. Blumenschein, S.~Cutler, J.~A.~R. Zepeda, N.~D. Naclerio,
  H.~El-Hussieny, U.~Mehmood, J.~Ryu, E.~W. Hawkes, and A.~M. Okamura, ``Vine
  robots: Design, teleoperation, and deployment for navigation and
  exploration,'' \emph{IEEE Robotics and Automation Magazine}, 2020,
  doi:10.1109/MRA.2019.2947538.

\bibitem{gonzalez2010toward}
J.~Gonzalez-Gomez, J.~Gonzalez-Quijano, H.~Zhang, and M.~Abderrahim, ``Toward
  the sense of touch in snake modular robots for search and rescue
  operations,'' in \emph{Proc. ICRA 2010 Workshop “Modular Robots: State of
  the Art}, 2010, pp. 63--68.

\bibitem{searle2013optical}
T.~C. Searle, K.~Althoefer, L.~Seneviratne, and H.~Liu, ``An optical curvature
  sensor for flexible manipulators,'' in \emph{2013 IEEE International
  Conference on Robotics and Automation}.\hskip 1em plus 0.5em minus
  0.4em\relax IEEE, 2013, pp. 4415--4420.

\bibitem{shahmiri2020sharc}
F.~Shahmiri and P.~H. Dietz, ``Sharc: A geometric technique for
  multi-bend/shape sensing,'' in \emph{Proceedings of the 2020 CHI Conference
  on Human Factors in Computing Systems}, 2020, pp. 1--12.

\bibitem{truby2020distributed}
R.~L. Truby, C.~Della~Santina, and D.~Rus, ``Distributed proprioception of 3d
  configuration in soft, sensorized robots via deep learning,'' \emph{IEEE
  Robotics and Automation Letters}, vol.~5, no.~2, pp. 3299--3306, 2020.

\bibitem{tapia2020makesense}
J.~Tapia, E.~Knoop, M.~Mutn{\`y}, M.~A. Otaduy, and M.~B{\"a}cher, ``Makesense:
  Automated sensor design for proprioceptive soft robots,'' \emph{Soft
  robotics}, vol.~7, no.~3, pp. 332--345, 2020.

\bibitem{zou2020design}
S.~Zou, Y.~Lv, Y.~Man, and W.~Han, ``Design and implement of shape detection
  for the soft manipulator,'' in \emph{2020 39th Chinese Control Conference
  (CCC)}.\hskip 1em plus 0.5em minus 0.4em\relax IEEE, 2020, pp. 3972--3977.

\bibitem{wallin20203d}
T.~J. Wallin, L.-E. Simonsen, W.~Pan, K.~Wang, E.~Giannelis, R.~F. Shepherd,
  and Y.~Meng{\"u}{\c{c}}, ``3d printable tough silicone double networks,''
  \emph{Nature communications}, vol.~11, no.~1, pp. 1--10, 2020.

\bibitem{pcbdesign}
``{MultiCircuitBoards:} general design rules for flexible pcbs,''
  \url{https://www.multi-circuit-boards.eu/en/pcb-design-aid/flex-rigid-flex.html},
  accessed: 2021-02-24.

\bibitem{smoothon}
``{VytaFlex 60 Product Information},'' \url{https://www.smooth-on.com/},
  accessed: 2021-02-24.

\bibitem{polymerdatabase}
``{PolymerDatabase},'' \url{http://www.polymerdatabase.com}, accessed:
  2021-02-24.

\end{thebibliography}

\end{document}